\documentclass{article}

\usepackage[preprint]{neurips_2024}
\setcitestyle{numbers}
\usepackage{xfrac}

 \usepackage{graphicx}
\usepackage[utf8]{inputenc} 
\usepackage[T1]{fontenc}    
\usepackage{hyperref}       
\usepackage{url}            
\usepackage{booktabs}       
\usepackage{amsfonts}       
\usepackage{nicefrac}       
\usepackage{microtype}      
\usepackage{algorithm}
\usepackage[noend]{algorithmic}
\usepackage{amsthm,amsmath,amssymb}
\usepackage{subcaption}
\usepackage[dvipsnames]{xcolor}

\title{WASH: Train your Ensemble with Communication-\\ Efficient Weight Shuffling, then Average}

\author{%
  Louis Fournier \\
  ISIR - Sorbonne Université \\
  Paris - France \\
  \texttt{louis.fournier@isir.upmc.fr} \\
  \And
Adel Nabli\\
  Mila, Concordia University\\
  Sorbonne Université \\
  \texttt{adel.nabli@sorbonne-universite.fr} \\
  \And
Masih Aminbeidokhti\\
  École de technologie supérieure \\ 
  Montréal - Québec 
  \And
Marco Pedersoli\\
  École de technologie supérieure\\ 
  Montréal - Québec 
  \And
   Eugene Belilovsky \\
   Concordia University\\
  Mila – Quebec AI Institute  \\
  Montréal - Québec 
  \And
  Edouard Oyallon\\
  Center for Computational Mathematics\\
  Flatiron Institute\\
  New York, USA
}

\begin{document}

\maketitle

\begin{abstract}
    The performance of deep neural networks is enhanced by ensemble methods, which average the output of several models. However, this comes at an increased cost at inference. Weight averaging methods aim at balancing the generalization of ensembling and the inference speed of a single model by averaging the parameters of an ensemble of models. Yet, naive averaging results in poor performance as models converge to different loss basins, and aligning the models to improve the performance of the average is challenging. Alternatively, inspired by distributed training, methods like DART and PAPA have been proposed to train several models in parallel such that they will end up in the same basin, resulting in good averaging accuracy. However, these methods either compromise ensembling accuracy or demand significant communication between models during training. In this paper, we introduce WASH, a novel distributed method for training model ensembles for weight averaging that achieves state-of-the-art image classification accuracy. WASH maintains models within the same basin by randomly shuffling a small percentage of weights during training, resulting in diverse models and lower communication costs compared to standard parameter averaging methods.  

\end{abstract}


\section{Introduction}
In order to enhance the accuracy of a given class of models, aggregating the answers of multiple instances trained in parallel can be done via model \emph{ensembling}. This can lead to significant improvements in modern deep learning models \citep{fort2019deep}, increasing generalization ability. However, this comes at the cost of evaluating multiple instances of a given model at inference. This increases both the required memory and computations, resources which can be critical for on-device inference \cite{menghani2023efficient}. To resolve this issue, the population of models can be fused into a single model to obtain both the generalization improvements of ensembling and the inference cost of a single model. Since independent models can be linearly connectable \cite{frankle2020linear}, a simple technique is to average the weights of the different models to obtain a fused model \cite{pmlr-v162-wortsman22a}.




There are however limits to this method. For models that are too dissimilar, the performance of the averaged model may not be better than chance \cite{izmailov2019averaging}. To mitigate this, the ensemble can either use a pre-trained network as a starting point \cite{NEURIPS2020_0607f4c7} or ensure that models share part of their optimization path \cite{frankle2020linear}. 
Nevertheless, diminishing too much the ensemble diversity comes at the cost of its performance (see Fig. 6 of \cite{fort2019deep}), revealing a tradeoff between model diversity and weight averageability. 
Inspired by distributed training, techniques like DART \cite{2302.14685_dart} and PAPA \cite{2304.03094_papa} have been proposed to train a population of models in parallel on heterogeneous data while communicating to balance this tradeoff. DART, similarly to LocalSGD \cite{stich2019localsgd}, averages all the models regularly to avoid models diverging. PAPA controls the diversity of the models more finely, by pushing them towards the averaged parameters using an Exponential Moving Average (EMA) like EASGD \cite{EASGD2015}, achieving better performances. Notably, they show that training a population in such a way results in models that generalize better than a single model trained with the same compute as the entire population, demonstrating the potential of these distributed approaches. However, existing methods require a regular computation of the average model using an all-reduce operation, either to remove periodically any diversity in the population \cite{2302.14685_dart} or in the case of PAPA, to compute an EMA of the average. This results in a high communication cost during the parallel training of the population of models \cite{pati2023computation}, hindering the scalability of these approaches as the population size increases \cite{ortiz2021tradeoffslocalsgd}.

\begin{figure}[t]
    \centering
    \includegraphics[width=0.9\linewidth]{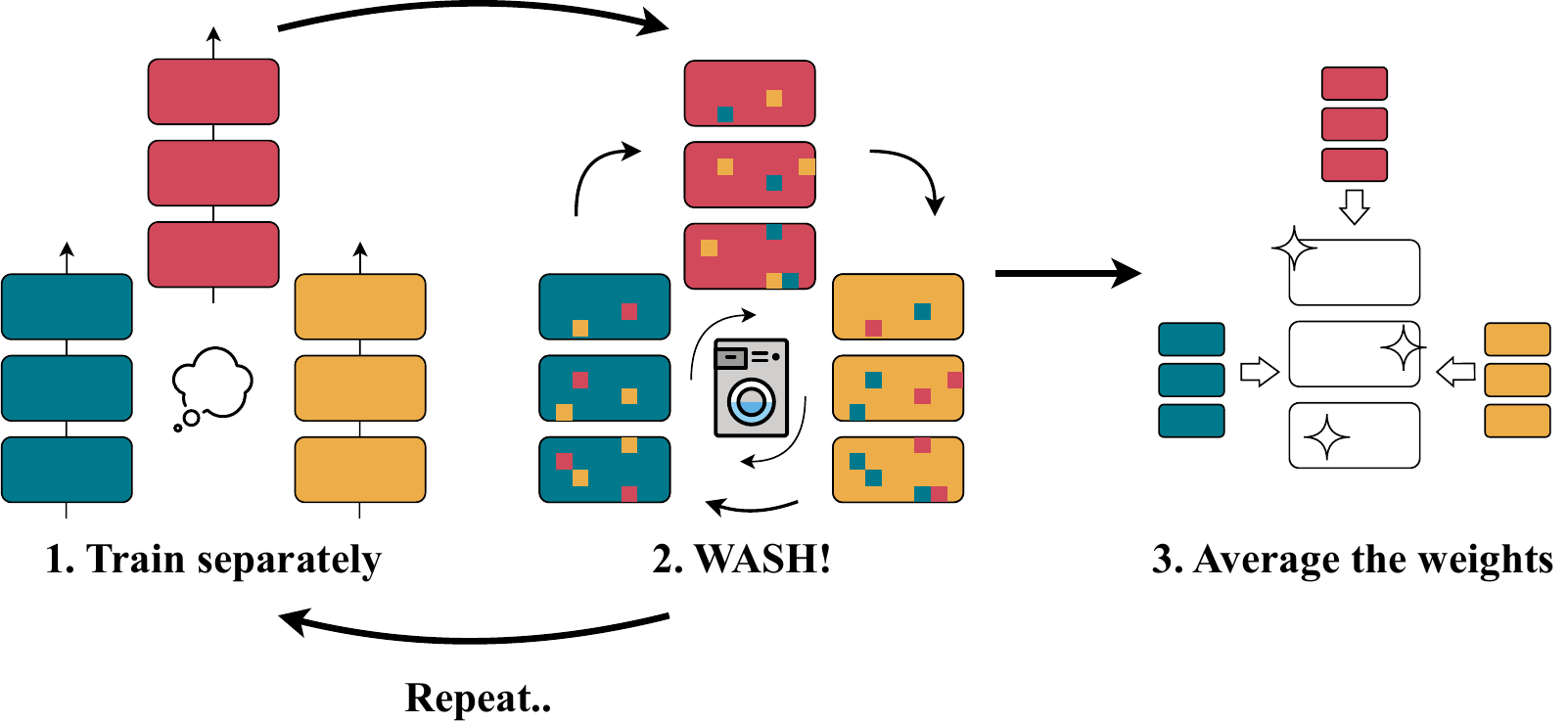}
    \caption{\textbf{Representation of training with WASH}. A population of models is being trained separately. \textbf{(1)} After each training step, \textbf{(2)} a small percentage of parameters are permuted between models. \textbf{(3)} At the end of the training, the model weights are averaged, resulting in a high-performance model.}
    \label{fig:representation}
\end{figure}

We propose in this paper a novel distributed method to train a population of models in parallel while keeping their weights within the same basin. It requires a fraction of the communication cost of PAPA but displays greater model diversity during training, increasing the final averaging accuracy. Our main idea is to shuffle parameters between models during training, forcing them to learn using the others' parameters. We refer to 'parameter shuffling' as the following idea. A permutation is randomly chosen, and models will communicate peer-to-peer their parameters following the permutation. 
The use of a permutation is distinct from the notion of weight permutation of \cite{2209.04836_git_rebasin} which is inside one model. 
We denote our method, which achieves \textbf{W}eight \textbf{A}veraging using parameter \textbf{SH}uffling, as  \textbf{WASH}, and represent it schematically in Fig. \ref{fig:representation}.




\paragraph{Contributions.} Our work makes the following contributions: \textbf{(1)} We propose a novel method for the training of a population of models that can be weight averaged, which we refer to as WASH (\textbf{W}eight \textbf{A}veraging using parameter \textbf{SH}uffling). 
By shuffling a small number of parameters between models during training, the resulting population can be weight-averaged into a high-performance model for a fraction of the communication volume of methods such as PAPA.   
\textbf{(2)} We find that WASH provides state-of-the-art results on image classification tasks, resulting in models with performances at the level of ensembling methods while only requiring a single network at inference time. \textbf{(3)} We provide experiments to better understand the improvement provided by WASH, notably on how WASH reduces the distance between models in the population implicitly while maintaining diversity. \textbf{(4)} We perform different ablations on our method, showing the impact of the shuffling. 
\textbf{(5)} Our code is made available at 
\hyperlink{https://github.com/fournierlouis/WASH}{github.com/fournierlouis/WASH}.

\section{Related work}

\paragraph{Ensemble and weight averaging.}
By combining predictions from multiple models, ensemble methods significantly improve a predictive system's ability to make accurate generalizations \cite{dietterich2000ensemble, lakshminarayanan2017simple}, while reducing the variance of estimator \cite{Breiman1996}. 
This variance reduction is especially effective when errors are uncorrelated and models show diversity, meaning they do not fail simultaneously on the same instances \cite{gontijo2021no, fort2019deep}. However, ensembles require additional passes through each model for inference, leading to increased computational costs. This cost can become prohibitive with a large number of models. As a remedy under some constraints, models can be averaged together to remove the computational burden during inference. Averaging the weights of models was first explored in 
simple linear \cite{lakshminarayanan2018linear} and convex scenarios \cite{polyak1992acceleration, bottou2018optimization}. In deep learning, 
\cite{izmailov2019averaging} establish that weight averaging is a first-order approximation of the ensemble when models are close in the weight space. Notably, a simple averaging of multiple points along the SGD trajectory leads to better generalization. 
Following mode connectivity \cite{NEURIPS2018_be3087e7_fge, frankle2020linear} and the observation that many optima of independent models are connectable, \cite{benton2021loss, 2102.10472_wortsman_learnlinesubspaces} propose
learning simplexes in parameter space with a regularization penalty to encourage diversity in weight space, and \cite{wortsman2022robust, 2205.09739} propose to train multiple model branches with different last-layer initialization and hyperparameters concurrently. These models are later averaged to enhance generalization and reduce the inference cost. However, for these models to be amenable to weight averaging, they need to begin with the same pre-trained initialization \cite{NEURIPS2020_0607f4c7} which can diminish diversity among the models. To alleviate this issue, neuron alignment techniques \cite{singh2023model, 2209.04836_git_rebasin, 2212.12042,horoi2024harmony} match the units of multiple networks to make them amenable to weight averaging, however, they rarely work in practical scenarios \cite{jordan2023repair}, often obtaining performance below the individual models. DART \cite{2302.14685_dart} and Branch-Train-Merge (BTM) \cite{li2022branchtrainmerge} propose a three-phase training pipeline. The process begins with an initial shared training phase, followed by the parallelized training of multiple models, each diversified through different data domains or different data augmentations. Finally, these models are merged into a single model. They find that iterative refinement of the last 2 stages enhances the overall optimization trajectory and improves generalization. To enhance diversity among models, PAPA \cite{2304.03094_papa} proposes to rather gradually adjust model weights towards the population average throughout the training process, beginning from random initialization. However, these approaches can result in substantial communication costs during training. Conversely, WASH tackles high communication costs by permuting only a small fraction of parameters among models during training, while ensuring the branches remain amenable to weight averaging at the end.



\paragraph{Distributed and federated learning.}
In distributed training of deep learning models, the trade-off between communications and model performance is a core concern \cite{ortiz2021tradeoffslocalsgd, Konecn2016FederatedOD}, and finding methods efficiently alleviating some of its communication cost is a recurrent theme in different areas of research \cite{wang2023zeropp, fournier2024cyclic}. For instance, communication overhead being a key concern of decentralized optimization, it has been shown in this literature that to train models in a data-parallel setting with a limited communication budget, a key metric to observe is the average distance to the consensus \cite{kong2021consensus, shi2014extra, tang2018d2, Wang_2021, nabli2023acid}. The techniques discussed earlier to train a population of models for weight averaging are similar to methods in the LocalSGD \cite{stich2019localsgd, Lin2020Dont} and Federated Learning \cite{pmlr-v54-mcmahan17a, Konecn2016FederatedOD, Fedprox2019} literature. The training in DART and BTM is similar to the LocalSGD training, where models are averaged regularly after several steps of computations. PAPA, which uses an EMA of the averaged model to gradually move the models towards consensus, is similar to methods like EASGD \cite{EASGD2015} or SlowMo \cite{Wang2020SlowMo}. Only averaging a population at the end of training like in BTM was also proposed for LocalSGD \cite{2106.04759_localsgdoneshotavg}, and cross-gradient aggregation \cite{pmlr-v139-esfandiari21a_crossgradient} can be seen as a way to shuffle gradients locally. Federated learning also uses techniques discussed previously for model merging \cite{2002.06440, pmlr-v97-yurochkin19a, chen2023fedsoup}. Finally, our method can be thought of as training a global model where each local model picks randomly from a subset of parameters when shuffled. This can be linked to Bayesian learning \cite{1506.02142_dropoutbayesian} in particular for federated learning \cite{2109.15258_feddropout, 2210.16105_asyncdropout}, or to federated subnetwork training \cite{pmlr-v180-dun22a_resist_layerwise, 2306.16484_richtarik_subnetwork}.

\begin{algorithm}[t]
\caption{Training with WASH}
\label{alg:wash_training}
\begin{algorithmic}[1]
\STATE \textbf{Input:} Datasets $D_i$, number of models $N$, initial parameters $\theta_0$, training steps $T$, number of layers $L$, base probability $p$
\STATE Initialize parameters $(\theta_n)_n \leftarrow \theta_0$ and optimizers $\textsc{opt}_i$
\FOR{$t = 1$ to $T$}
 \STATE  \textit{\# Training step}
    \FOR{$n = 1$ to $N$, in parallel}
    \STATE $(x_n, y_n) \leftarrow D_n$   \textit{\ \ \ \ \ \ \ \ \ \ \ \ \ \ \ \ \ \ \# Sample data} 
    \STATE $\theta_n \leftarrow \textsc{opt}_n(x_n, y_n, \theta_n)$\textit{\ \ \ \ \  \# Update the model $n$} 
\ENDFOR
 \STATE  \textit{\# Shuffling step}
    \FOR{layer $l = 0$ to $L-1$}
    \FOR{parameter $\theta^i $ in layer $l$}
    \STATE \textbf{With} probability $p(1-\frac{l}{L-1})$, 
    \STATE \hspace{\algorithmicindent} $\pi_i \leftarrow$ Random permutation
    \STATE \hspace{\algorithmicindent} $(\theta^i_n)_n \leftarrow (\theta^i_{\pi_i(n)})_n$ \textit{\ \ \ \# Send and permute the parameter} 
\ENDFOR\ENDFOR\ENDFOR
\STATE \textbf{Output:} the averaged model $\frac{1}{N} \sum_{n=1}^{N} \theta_n$
\end{algorithmic}
\end{algorithm}

\section{Parameter shuffling in an ensemble for weight averaging}

\paragraph{Motivation of our training procedure.} 
We aim to balance the benefits of model ensembling with the computational efficiency of using a single model at inference via weight averaging. In other words, our objective is to produce a single model resulting from the ensembling. A set of $N$ model parameters $\{\theta_n\}_{n \leq N} \subset \mathbb R^d$  are trained in parallel on the same dataset, with a different data order and possibly different data augmentations and regularizations. To avoid divergence among the models, PAPA applies an EMA every $T$ training steps and produces the following update

\begin{equation}
    \tilde{\theta}_n \leftarrow \alpha \theta_n + (1-\alpha) \bar{\theta}\,, \label{eq:papa}
\end{equation}

where $\bar{\theta} \triangleq \frac{1}{N} \sum_{n=1}^N \theta_n$ represents the average of the model weights, also referred to as the \textit{consensus}, and $\alpha\in]0,1[$ is weighted depending on the learning rate. Despite its benefits, this method has drawbacks, including the need for synchronized global communication across all models, which can be inefficient, and the potential reduction in model diversity due to the consensus constraint, which may reduce model expressivity. Indeed, we observe that after each update
\begin{equation}
    \sum_n \lVert \tilde{\theta}_n  -  \bar{\theta} \rVert^2 = \alpha^2 \sum_n\lVert \theta_n  -  \bar{\theta} \rVert^2 <  \sum_n \lVert {\theta}_n  -  \bar{\theta} \rVert^2 \,, \label{eq:consenspapa}
\end{equation}
which shows that the EMA step of methods like PAPA directly reduces the distance of the models to the consensus, hindering their diversity. 

\paragraph{Proposed method: WASH.} To address these challenges, we propose the following stochastic parameter shuffling step instead of the EMA, defined for any individual parameter $\theta^i_n \in \mathbb R$ of a model $\theta_n=[\theta^i_n]_{i=1}^d$ by 


\begin{equation}
    \hat{\theta}^i_n \leftarrow 
    \begin{cases} 
    \theta^i_{\pi_i(n)} & \text{with probability } p, \\
    \theta^i_n & \text{otherwise}, 
    \end{cases} \label{eq:wash}
\end{equation}

where $\pi_i$ denotes a random permutation of the indices $\{1, ..., N\}$, chosen uniformly at each iteration for each parameter index $i \in \{1,...,d\}$, and independently from the Bernoulli variable of Eq. \eqref{eq:wash}.  Notably, this parameter shuffling reduces in expectation to 
\begin{equation}
    \mathbb{E}[\hat{\theta}_n] =  (1-p) \theta_n + p \bar{\theta} \,. \label{eq:expwash}
\end{equation}
Thus, WASH aligns, in expectation, with the EMA of Eq.~\eqref{eq:papa} for $p=(1-\alpha)$. The expected number of parameters communicated by each model at each step is thus $p \times d$ while for PAPA, each model communicating all of its parameters every $T$ steps, this amounts to $ \frac d T$. Thus, $p\ll \frac 1T$ results in a significantly reduced communication overhead favorable to WASH. However, the model diversity is higher, as WASH preserves the consensus distance, as shown by
\begin{equation}
     \sum_n\lVert \hat{\theta}_n -  \bar{\theta} \rVert^2  = \sum_n \sum_i( \hat{\theta}^i_n -  \bar{\theta}^i )^2 = \sum_i \sum_n( {\theta}^i_n -  \bar{\theta}^i )^2 =\sum_n \lVert {\theta}_n -  \bar{\theta} \rVert^2 \,. \label{eq:consenswash}
\end{equation}

\paragraph{Layer-wise adaptation via WASH.} Recognizing that different network layers may require varying levels of adaptation due to their roles and dynamics, we introduce a layer-specific probability adjustment. Assuming $L$ layers in the network, we set for each layer $l$ (where $0 \leq l < L$)

\begin{equation}
    p_l = p \left(1-\frac{l}{L-1}\right)\,,
\end{equation}

where $p$ is a base probability. In other words, the parameters of the first layer have a shuffling probability of $p$, while the final layer's parameters are never shuffled. This adaptation ensures that deeper layers, which are typically slower to train and more sensitive to the input features, undergo fewer permutations than the more generalizable early layers. This strategy not only preserves the specificity required by the initial layers but also further halves the overall communication overhead. 
\paragraph{Full procedure.} 
Alg. \ref{alg:wash_training} presents the training of a population of $N$ models using WASH. Starting from the same initialization, our training procedure alternates between local gradient computations and shuffling communications. At inference, we simply average the weights of the models, obtaining a single model with parameters $\bar\theta$. Note that techniques like REPAIR \cite{jordan2023repair} or activation alignment \cite{2209.04836_git_rebasin} could be incorporated to improve the alignment of the models, but we found them to be unnecessary to obtain high accuracy and kept our evaluation framework minimal for the sake of simplicity.

\section{Experiments}


\paragraph{Training methods.} We showcase the capacities of WASH for training a population of neural networks on standard image classification tasks. As a Baseline, we consider a population trained separately, with each model working on a different dataset order and different data augmentations and regularization (if they are used). This is the same baseline as \cite{2304.03094_papa}, only starting from the same initialization, but we found that this change brought no significant impact on performance. 
We also compare WASH to PAPA \cite{2304.03094_papa} on the same tasks (with PAPA however using models with a different initialization), to show our improvement despite requiring a fraction of the communication cost. We do not provide comparisons to DART \cite{2302.14685_dart} or the variants of PAPA as their performances are generally inferior~\cite{2304.03094_papa}. We also propose a variant of WASH named WASH+Opt that also permutes the optimizer state associated with the parameter shuffled (in our case, the momentum of SGD), doubling the volume of communications. We do not permute or recompute the running statistics of BatchNorm layers for simplicity.

\begin{table}[t]
\centering
\caption{\textbf{Communication volume and inference costs} of four training techniques. The baseline Ensemble is trained separately but requires a linearly increasing inference cost. In our experiments, we fix the base probability of WASH and WASH+Opt to be equal to $0.001$ or $0.05$ when training on CIFAR-10/100 or ImageNet, resulting in a reduction of communication volume over PAPA. }
\label{tab:costs}
\begin{tabular}{llll}
\toprule
&   \multicolumn{2}{c}{\textbf{Communication volume}} & \\
\textbf{Technique} & \textbf{CIFAR-10/100} & \textbf{ImageNet} & \textbf{Inference cost}   \\ \midrule
Ensemble  & $ 0$ & $ 0$  & $ N$\\ 
PAPA   & $ 1$ & $ 1$ & $ 1$ \\
WASH  & $ \sfrac{1}{200}$ & $ \sfrac{1}{4}$& $ 1$  \\
WASH+Opt & $ \sfrac{1}{100}$ & $ \sfrac{1}{2}$& $ 1$  \\ \bottomrule
\end{tabular}
\end{table}

\paragraph{Communication cost.}
Training with PAPA requires computing an all-reduce operation on all of the models' parameters every $T=10$ training steps. In comparison, WASH requires, in expectation, a shuffling of $p/2$ of the parameters of the models at every training step. Thus, by keeping a base probability $p \leq 0.2$, WASH results in a more communication-efficient training. In practice, $p$ will be equal in our experiments to $0.001$ or $0.05$, ensuring a communication volume reduction of $200$ or $4$.

\paragraph{Evaluation strategy.} After training, the population of models obtained can be evaluated in three separate ways. As a baseline, the performance of the population can be evaluated as an Ensemble, averaging the predictions of the models. The parameters of the models can be averaged to obtain a single model, which we refer to as Averaged. This is equivalent to UniformSoup in \cite{pmlr-v162-wortsman22a} or AvgSoup in \cite{2304.03094_papa} for example. More elaborate averaging methods have been proposed, such as GreedySoup \cite{pmlr-v162-wortsman22a}, which averages an increasing number of models (in order of validation accuracy) until the averaging does not improve accuracy. We report the accuracy of the Ensemble and Averaged model for all training techniques, as well as the GreedySoup accuracy of the Baseline. Like \cite{2304.03094_papa}, we find that the GreedySoup accuracy corresponds to the accuracy of a single model for the Baseline and that the Averaged model accuracy outperforms the GreedySoup model for the other techniques, and thus chose not to report it. We summarize in Tab.~\ref{tab:costs} the communication volume and inference cost necessary for training a separate Ensemble of models, or training with PAPA, WASH, or WASH+Opt.


\begin{table}[t]
\centering
\caption{\textbf{Ensemble and Averaged Model accuracy for a heterogeneous population of models; trained with varying data augmentations and regularizations.} We compare models trained separately (Baseline), with PAPA, or our method WASH and its variant WASH+Opt. We also report the GreedySoup accuracy for the Baseline models.  The best Ensemble (black) and Averaged (blue) accuracy are reported in bold. Except on CIFAR-10, WASH and in particular WASH+Opt provide the best performance for the final Averaged Model, with performances comparable to the Ensemble of models for a fraction of the inference cost.}
\label{tab:main_results}
\resizebox{\textwidth}{!}{
\begin{tabular}{ll|lll|ll|ll|ll}
\toprule
\multicolumn{2}{c|}{\textbf{Method} }                   &      \multicolumn{3}{|c|}{\textbf{Baseline (trained separately)} }                 &  \multicolumn{2}{|c}{\textbf{PAPA} }  &  \multicolumn{2}{|c|}{\textbf{WASH (ours)} }   &  \multicolumn{2}{|c}{\textbf{WASH+Opt (ours)} }         \\ 
\textbf{Config} &  \textbf{\#N}                         & Ensemble  & Averaged  & GreedySoup  & Ensemble  & Averaged  & Ensemble  & Averaged  & Ensemble  & Averaged \\ \midrule\midrule
\multicolumn{2}{l|}{\textbf{CIFAR-10}}   &             &              &                &           &              &         &              &              &              \\ \midrule
\textbf{VGG-16}     & 3              & 95.98$\pm$.42 &	10.00$\pm$.00 &	95.26$\pm$.05       & \textbf{96.12$\pm$.34}             & \textcolor{Blue}{\textbf{96.13$\pm$.24}}    & 95.89$\pm$.23            & 95.97$\pm$.24                         & 95.91$\pm$.36             & 95.85$\pm$.27            \\
                    & 5              &\textbf{96.28$\pm$.40} & 	10.00$\pm$.00 &	95.42$\pm$.10           & 96.24$\pm$.17             &  \textcolor{Blue}{\textbf{96.21$\pm$.13}}    & 96.15$\pm$.10             &  \textcolor{Blue}{\textbf{96.20$\pm$.10}} & 96.00$\pm$.21             & 96.04$\pm$.14            \\
                    & 10             & \textbf{96.47$\pm$.07} &  10.00$\pm$.00  &  95.39$\pm$.24  & 96.32$\pm$.13             &  \textcolor{Blue}{\textbf{96.31$\pm$.13}}    & 96.27$\pm$.10             & 96.18$\pm$.13                         & 96.14$\pm$.08             & 96.20$\pm$.05            \\ \midrule
\textbf{ResNet18}   & 3              & 97.15$\pm$.28                     & 10.17$\pm$.29                     & 96.62$\pm$.38                      & \textbf{97.33$\pm$.05}             &  \textcolor{Blue}{\textbf{97.24$\pm$.05}}             & 97.21$\pm$.19            & 97.19$\pm$.17                         & 97.22$\pm$.07 &	 \textcolor{Blue}{\textbf{97.25$\pm$.14}}     \\
                    & 5              & \textbf{97.33$\pm$.08}                     & 10.09$\pm$.16                    & 96.61$\pm$.03                      & \textbf{97.35$\pm$.12}             &  \textcolor{Blue}{\textbf{97.31$\pm$.06}}             & 97.21$\pm$.10             & 97.25$\pm$.12                & 97.18$\pm$.09             & 97.16$\pm$.07            \\
                    & 10             & \textbf{97.59$\pm$.01}                     & 9.26$\pm$1.28                      & 96.79$\pm$.14                      &   97.39$\pm$.13 &	 \textcolor{Blue}{\textbf{97.34$\pm$.06}}          & 97.30$\pm$.10             & 97.28$\pm$.04                         & 97.20$\pm$.13             & 97.16$\pm$.13            \\ \midrule
\multicolumn{2}{l|}{\textbf{CIFAR-100}}             & \multicolumn{1}{l}{}                 & \multicolumn{1}{l}{}                 & \multicolumn{1}{l|}{}                 & \multicolumn{1}{l}{}         & \multicolumn{1}{l|}{}        & \multicolumn{1}{l}{}         & \multicolumn{1}{l|}{}                    & \multicolumn{1}{l}{}         & \multicolumn{1}{l}{}        \\ \midrule
\textbf{VGG-16}     & 3              & \textbf{80.36$\pm$.1}5 & 1.00$\pm$.00 & 77.92$\pm$.22                & 78.89$\pm$.10             & 78.77$\pm$.16             & 79.10$\pm$.88             & 79.05$\pm$.68                         & 79.15$\pm$.61             &  \textcolor{Blue}{\textbf{79.15$\pm$.41}}   \\
                    & 5              & \textbf{81.32$\pm$.56} &	1.00$\pm$.00 & 77.81$\pm$.25                & 79.51$\pm$.38             & 79.24$\pm$.43             & 79.65$\pm$.27             & 79.39$\pm$.21                         & 79.75$\pm$.21             &  \textcolor{Blue}{\textbf{79.71$\pm$.20}}   \\
                    & 10             & \textbf{82.24$\pm$.15}            & 1.00$\pm$.00       & 77.83$\pm$.65                      & 79.95$\pm$.11             & 79.64$\pm$.13             & 80.05$\pm$.18             & 79.70$\pm$.25                         & 80.03$\pm$.11             &  \textcolor{Blue}{\textbf{79.76$\pm$.13}}   \\ \midrule
\textbf{ResNet18}   & 3              & \textbf{82.84$\pm$.48}            & 1.00$\pm$.01                      & 80.06$\pm$1.5                      & 81.58$\pm$.12             & 81.53$\pm$.13             & 81.91$\pm$.34             & 81.90$\pm$.36                         & 81.99$\pm$.06             & \textcolor{Blue}{\textbf{82.08$\pm$.09}}  \\
                    & 5              & \textbf{83.72$\pm$.49}           & 1.00$\pm$.00                      & 80.72$\pm$.52                      & 82.09$\pm$.30             & 82.01$\pm$.34             & 82.16$\pm$.42             & 81.97$\pm$.28                         & 82.35$\pm$.17             & \textcolor{Blue}{\textbf{82.17$\pm$.15}}   \\
                    & 10             & \textbf{84.18$\pm$.20}            & 1.00$\pm$.00                      & 80.61$\pm$.43                      & 82.32$\pm$.09 &	82.15$\pm$.14  & 82.43$\pm$.32             & \textcolor{Blue}{\textbf{82.31$\pm$.38}}                         & 82.42$\pm$.31             & 82.18$\pm$.22            \\ \midrule
 \textbf{ImageNet}  &      &              &                                                &              &              &                &              &              &              &              \\ \midrule
\textbf{ResNet50} & 3 & \textbf{76.16$\pm$.28} & 0.10$\pm$.00 & 74.15$\pm$.11            &   75.62$\pm$.15 & *      &  74.39$\pm$.14 & \textcolor{Blue}{\textbf{74.34$\pm$.18}}         &        74.30$\pm$.22 & 74.18$\pm$.26           \\
& 5         &   \textbf{76.68$\pm$.06} & 0.10$\pm$.00 & 74.47$\pm$.06         &       75.80$\pm$.21 &  *        &  74.63$\pm$.11 & \textcolor{Blue}{\textbf{74.59$\pm$.07}}  &    74.44$\pm$.21 & 74.39$\pm$.21       \\ \bottomrule
\end{tabular}
}
\end{table}

\subsection{Main experiments}

\paragraph{Experimental setup.} We showcase the performance of WASH for training neural networks on image classification tasks on the CIFAR-10, CIFAR-100 \cite{Krizhevsky2009LearningMLcifar10}, and ImageNet \cite{deng2009imagenet} datasets. We use the same training framework as \cite{2304.03094_papa} for a fair comparison. We train a population of $N$ models for $N \in \{3,5,10\}$, on the ResNet-18, 50 and VGG-16 architectures. 2\% of the training data is kept as validation for computing the GreedySoup. Like \cite{2304.03094_papa}, we consider one framework with heterogeneous models, learning with different data augmentations and regularizations, and one homogeneous setting with no data augmentations, where the only difference between the models' training is the dataset shuffling. Details are presented in the Appendix. Models are trained with SGD with momentum, a weight decay of $10^{-4}$, and a cosine annealing scheduler with starting and minimum learning rates $0.1$ and $10^{-4}$. For CIFAR-10/100, we train over 300 epochs with a batch size of 64, and 90 epochs with a batch size of 256 for ImageNet. For WASH and WASH-Opt we initialize the models with the same parameters and choose $p$ with cross-validation to be equal to $0.001$ or $0.05$ when training on CIFAR-10/100 or ImageNet. We do not require any alignment technique such as REPAIR.

\paragraph{Main results.} Tab. \ref{tab:main_results} and Tab. \ref{tab:main_results_noaug} correspond respectively to the heterogeneous and homogeneous settings. We report the test accuracies as the average of 3 runs for the Ensemble of models, the Averaged model, and the GreedySoup for the Baseline (equivalent to the best model). Consistent with the findings of \cite{2304.03094_papa}, we find that networks trained separately have a high Ensemble accuracy, but perform as random when averaged. On CIFAR-10/100, methods like PAPA and WASH result in lower Ensemble accuracy but almost no difference between the Ensemble and Averaged accuracies. 
In general, WASH and WASH+Opt outperform PAPA, despite requiring a lower communication volume. On ImageNet, our parallelization procedure resulted in a slightly lower Baseline accuracy and we were not able to reproduce PAPA's baseline due to our distributed constraint. The WASH Averaged model reaches a high accuracy, like previously. 
Both of our methods reduce the gap with accuracies of the baseline Ensemble, indicating that WASH hinders less the diversity of the population of models while maintaining weight averagability. However, a gap still remains, which may be inherent to models being in the same basin.  WASH and WASH+Opt have very similar results, with the simpler WASH being better on the homogeneous case and WASH+Opt being better on the heterogeneous case.

\begin{table}[t]
\centering
\caption{\textbf{Ensemble and Averaged Model accuracy for a homogeneous population of models.} We compare models trained separately (Baseline), with PAPA, or our methods WASH and WASH+Opt. The best Ensemble (black) and Averaged (blue) accuracy are reported in bold. We observe the same results in this setting, with WASH in particular reaching close to the Ensemble performance.}
\label{tab:main_results_noaug}
\resizebox{\textwidth}{!}{
\begin{tabular}{ll|lll|rr|rr|rr}
\midrule
\multicolumn{2}{c|}{\textbf{Method}} & \multicolumn{3}{c|}{\textbf{Baseline (trained separately)}}                                                                  & \multicolumn{2}{c|}{\textbf{PAPA}}                          & \multicolumn{2}{c|}{\textbf{WASH (ours)}}                   & \multicolumn{2}{c}{\textbf{WASH+Opt (ours)}}               \\
\textbf{Config}     & \textbf{\#N}   & Ensemble                                      & Averaged                              & GreedySoup                            & \multicolumn{1}{l}{Ensemble} & \multicolumn{1}{l|}{Averaged} & \multicolumn{1}{l}{Ensemble} & \multicolumn{1}{l|}{Averaged} & \multicolumn{1}{l}{Ensemble} & \multicolumn{1}{l}{Averaged} \\ \midrule
\multicolumn{2}{l|}{\textbf{CIFAR-10}}             &                                               &                                      &                                       & \multicolumn{1}{l}{}         & \multicolumn{1}{l|}{}        & \multicolumn{1}{l}{}         & \multicolumn{1}{l|}{}        & \multicolumn{1}{l}{}         & \multicolumn{1}{l}{}        \\ \midrule
\textbf{VGG-16}     & 3              & \textbf{94.93$\pm$.06} & 10.00$\pm$.00 & 93.60$\pm$.41                                 & 94.38$\pm$.14 &	94.34$\pm$.18   & 94.41$\pm$.23             & \textcolor{Blue}{\textbf{94.58$\pm$.17}}            & 94.45$\pm$.05             & 94.47$\pm$.02            \\
                    & 5              & \textbf{95.29$\pm$.05} &	10.00$\pm$.00 &	93.82$\pm$.30  & 94.55$\pm$.12             & 94.58$\pm$.12             & 94.72$\pm$.08             & \textcolor{Blue}{\textbf{94.70$\pm$.17}}             & 94.63$\pm$.11             & \textcolor{Blue}{\textbf{94.68$\pm$.14}}  \\
                    & 10             & \textbf{95.23$\pm$.06}                     & 10.00$\pm$.00                     & 93.82$\pm$.06                      & 94.79$\pm$.18             & \textcolor{Blue}{\textbf{94.78$\pm$.20}}             & 94.66$\pm$.03             & 94.54$\pm$.07             & 94.71$\pm$.07             & 94.61$\pm$.13   \\ \midrule
\textbf{ResNet18}   & 3              & \textbf{96.14$\pm$.10}                     & 10.00$\pm$.00                     & 95.42$\pm$.27     & 95.89$\pm$.04             & \textcolor{Blue}{\textbf{95.89$\pm$.06}}    & 95.77$\pm$.12             & 95.77$\pm$.17             & 95.85$\pm$.04             & \textcolor{Blue}{\textbf{95.87$\pm$.10}}   \\
                    & 5              & \textbf{96.19$\pm$.16} & 10.00$\pm$.00 & 95.31$\pm$.09 & 95.99$\pm$.08             & \textcolor{Blue}{\textbf{95.99$\pm$.08}}           & 95.96$\pm$.08             & \textcolor{Blue}{\textbf{95.98$\pm$.05}}    & 95.94$\pm$.12             & \textcolor{Blue}{\textbf{95.98$\pm$.12}}   \\
                    & 10             & \textbf{96.34$\pm$.02} & 10.00$\pm$.00 & 95.26$\pm$.11 & 96.10$\pm$.25             & \textcolor{Blue}{\textbf{96.11$\pm$.24}}    & 96.08$\pm$.07             & \textcolor{Blue}{\textbf{96.12$\pm$.09}}    & 96.07$\pm$.07             & 96.08$\pm$.14            \\ \midrule
\multicolumn{2}{l|}{\textbf{CIFAR-100}}               & \textbf{}                                     &                                      &                                       &         &        &  &   & & \\ \midrule
\textbf{VGG-16}     & 3              & \textbf{77.63$\pm$.24} & 1.00$\pm$.00  & 73.76$\pm$.35 & 75.10$\pm$.11             & 75.09$\pm$.16             & 76.30$\pm$.37             & \textcolor{Blue}{\textbf{76.04$\pm$.58}}    & 76.04$\pm$.03             & 75.96$\pm$.18            \\
                    & 5              & \textbf{78.52$\pm$.10} & 1.00$\pm$.00  & 73.76$\pm$.18 & 75.56$\pm$.16             & 75.55$\pm$.14             & 76.63$\pm$.27             & \textcolor{Blue}{\textbf{76.48$\pm$.23}}    & 76.64$\pm$.15             & 76.13$\pm$.18            \\
                    & 10             & \textbf{79.26$\pm$.06}                     & 1.00$\pm$.00                      & 73.99$\pm$.26                      & 76.24$\pm$.44             & 76.26$\pm$.43             & 77.06$\pm$.12             & \textcolor{Blue}{\textbf{76.43$\pm$.18}}    & 76.72$\pm$.15             & 75.94$\pm$.26            \\ \midrule
\textbf{ResNet18}   & 3              & \textbf{79.54$\pm$.17} &  1.00$\pm$.00  & 76.84$\pm$.54 & 77.83$\pm$.26             & 77.86$\pm$.30             & 78.90$\pm$.17             & \textcolor{Blue}{\textbf{78.76$\pm$.25}}    & 78.66$\pm$.08             & 78.56$\pm$.21            \\
                    & 5              & \textbf{80.11$\pm$.23} &  1.00$\pm$.00  & 76.83$\pm$.45 & 77.94$\pm$.16             & 77.92$\pm$.19             & 79.24$\pm$.32             & 79.09$\pm$.43             & 79.32$\pm$.19             & \textcolor{Blue}{\textbf{79.19$\pm$.15}}   \\
                    & 10             & \textbf{80.55$\pm$.13}                     & 1.00$\pm$.00                      & 76.80$\pm$.41                      & 78.40$\pm$.15             & 78.44$\pm$.22             & 79.65$\pm$.17             & \textcolor{Blue}{\textbf{79.43$\pm$.16}}    & 79.34$\pm$.34             & 79.19$\pm$.45            \\ \bottomrule
\end{tabular}
}
\end{table}

\begin{figure}[t]
    \centering
    \includegraphics[width=0.65\linewidth]{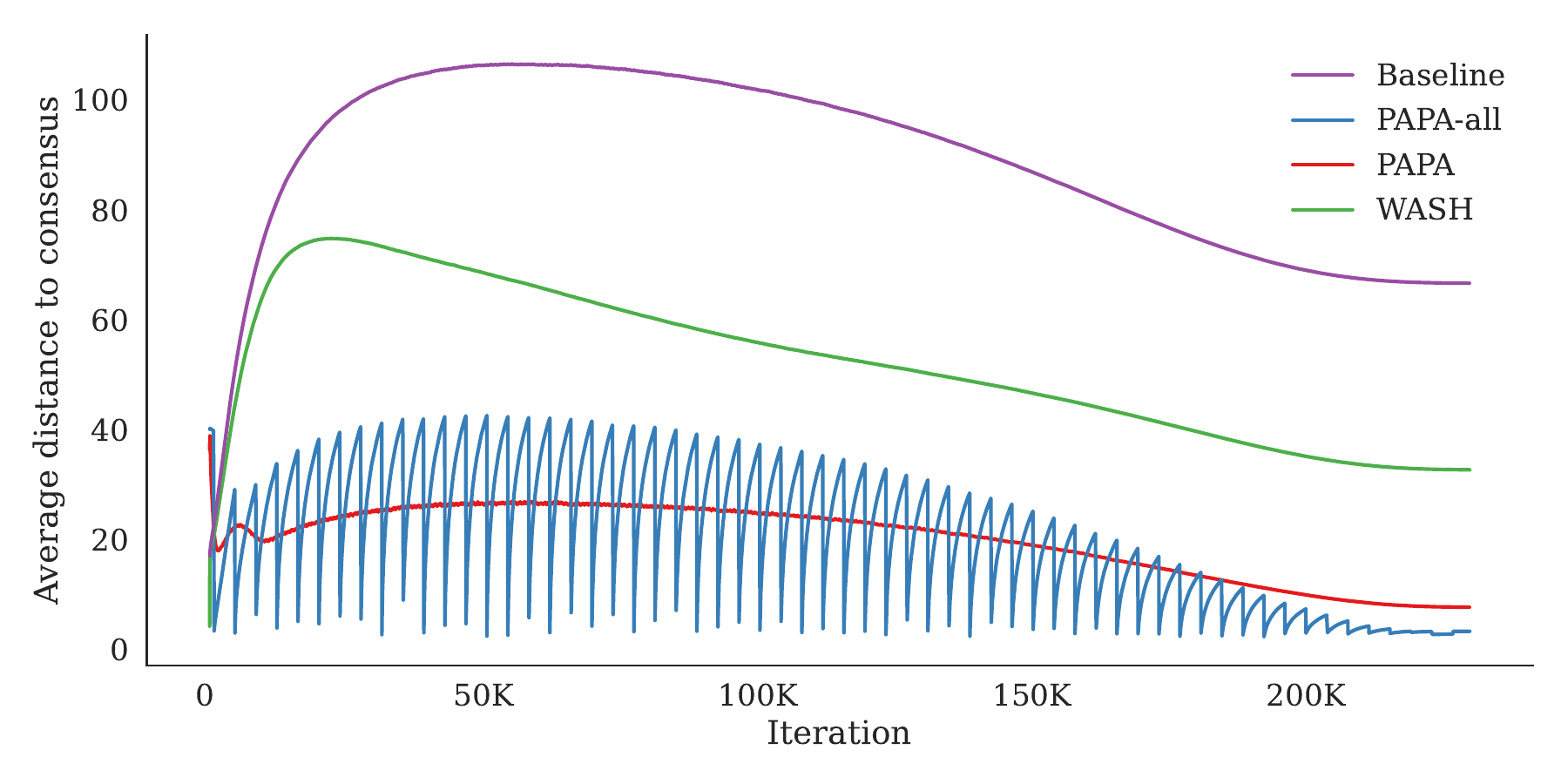}
    \caption{\textbf{Average distance to the consensus (i.e. the averaged model)} during training for a heterogeneous population of $5$ models trained on CIFAR-100, either separately, with PAPA, PAPA-all, or our method WASH. Starting at consensus, models initially diverge from each other before converging back during convergence, mainly due to weight decay. Models trained with WASH have a smaller distance to consensus than ones trained separately; allowing them to be averaged with no performance loss. 
    By training with PAPA-all (i.e. averaging to a single model every few epochs), models are not able to reach the same diversity as WASH between these averaging steps. Finally, the EMA of PAPA has a strong pulling effect towards consensus, resulting in a similar distance as PAPA-all. The jitter in the curve is due to the immediate distance reduction caused by the EMA steps.}
    \label{fig:consensus_base_papa}
\end{figure}

\begin{figure}[h]
    \centering
    \includegraphics[width=0.5\linewidth]{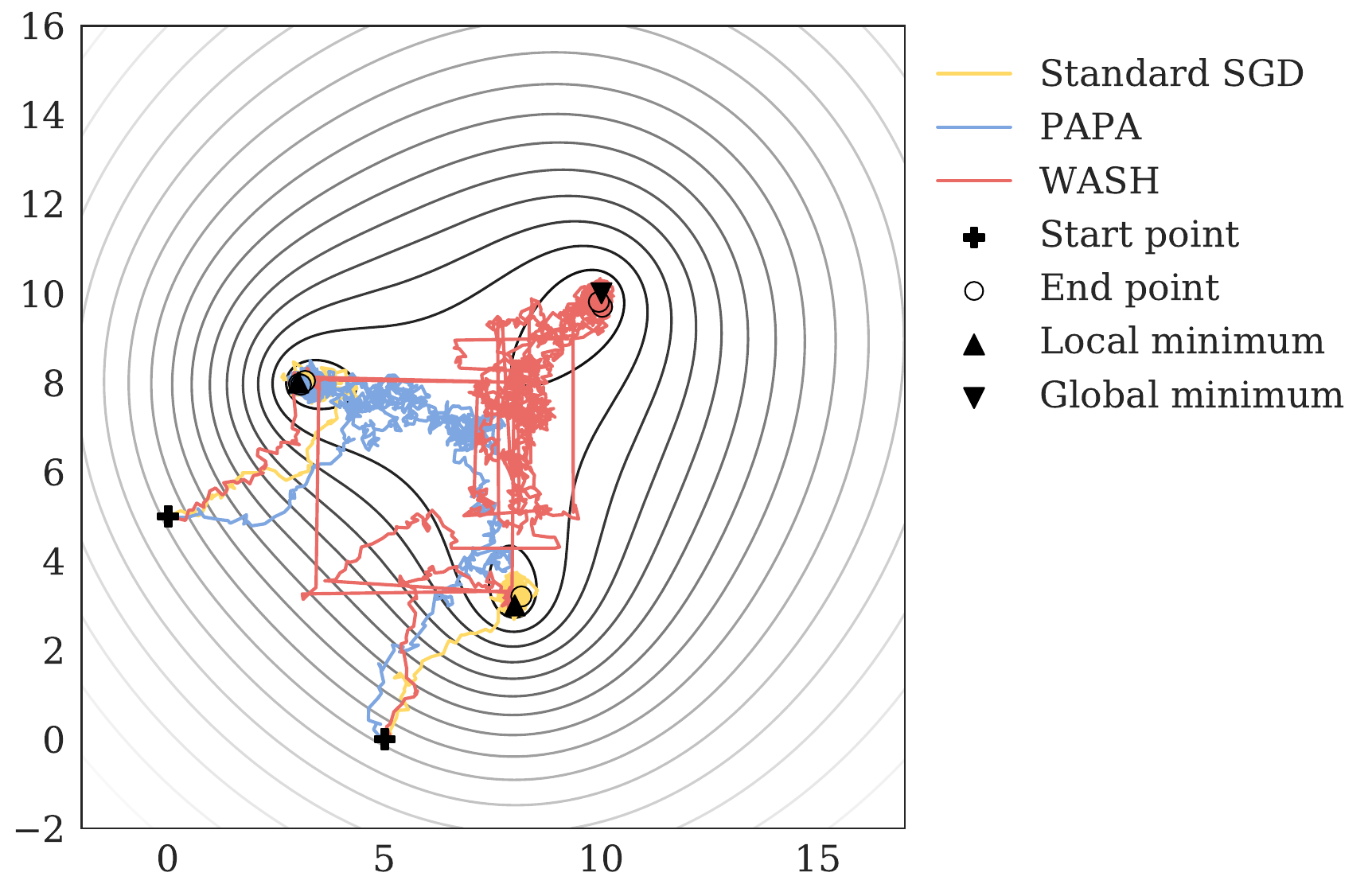}
    \caption{\textbf{2D optimization example.} We train 2 points with SGD on a simple loss function with 2 local and 1 global minima (upwards and downwards triangle). The two models are trained from two different starting points (plus signs). If the points are trained separately (yellow), they converge to their closest local minimum (yellow circles). By training with PAPA (blue), the points reach a consensus but then converge to one of the local minima (blue circles). By training with WASH (red), the shuffling (seen by the horizontal and vertical lines in the trajectory) allows more diversity in the optimization path, and the points both reach the global minimum (red circles).}
    \label{fig:2d_optim}
\end{figure}

\subsection{Why do shuffling parameters help?}

In this section, we propose to explain the improvement provided by our parameter shuffling over previous mechanisms such as BTM, DART, or PAPA, that focus on the averaging of parameters. We first show that models trained with WASH have a smaller distance to consensus than models trained separately. Then, we argue that despite this, WASH is a weak perturbation on the training of the models and that it incites diversity in the models.

\paragraph{Reducing distance to consensus.}

To better analyze the diversity of the models trained with WASH, we propose to report the distance of the models to the consensus (the averaged model) during training, as a proxy of the diversity metric. \cite{izmailov2019averaging, wortsman2022robust} showed that the difference between the Ensemble and the Averaged models depends on the distance between models. We present in Fig. \ref{fig:consensus_base_papa} the average distance of the models to the consensus, for models trained separately, with PAPA, PAPA-all, or with WASH. PAPA-all is a variant of PAPA functionally identical to DART. The idea is to average the weights every few epochs before letting the models diversify again. 
We observe that WASH results in a consistently lower distance to consensus than the baseline, despite explicitly leaving the distance to consensus unchanged during the shuffling step, and shuffling only a small number of parameters. Thus, the smaller distance at the end of the training explains why the averaging of the parameters does not cause a decrease in performance. By comparison, PAPA-all (i.e. DART) results in alternating phases where models diversify before being averaged, and we observe that the models are not able to reach the diversity of WASH. Similarly, the EMA of PAPA has a strong pulling effect 
and results in average in a similar diversity as PAPA-all. Thus, we find that models trained with WASH have a higher diversity than models trained with PAPA or PAPA-all, while being close enough that averaging them does not cause a loss of performance. More generally, we show in Fig.~\ref{fig:heatmaps} of the Appendix that various interpolations of models trained by WASH result in a similar performance, showcasing that they all reside in the same loss basin. 



\paragraph{Encouraging diversity.} WASH can be viewed as a weak perturbation on the models: the parameter shuffling affects more weakly the models than parameter averaging or the EMA of PAPA as only a few parameters are affected at a time, and the consensus distance is unaffected. Furthermore, the shuffling of parameters increases the diversity of the trajectories seen by the models. We showcase this with a toy example, by training jointly two points with SGD on a 2D loss function with 2 local minima and 1 global minimum, either training them separately, with PAPA, or with WASH. We represent the trajectories corresponding to each method in Fig.~\ref{fig:2d_optim}. Training separately the two points makes them converge into a separate local minimum (i.e. a different basin). Training with PAPA allows the two points to reach a consensus, however they converge together into a local minimum. In contrast, by training with WASH, we show that both points reach the global minimum, as the shuffling allows for a greater diversity of points to optimize with. We provide more details in the Appendix.

%

\subsection{Ablations}

We present in this section ablations to better understand the effect of the parameter shuffling, varying the layer-wise probability adaptation, the base probability value, and the shuffling period.

 \begin{figure}[t]
    \centering
    \includegraphics[width=0.72\linewidth]{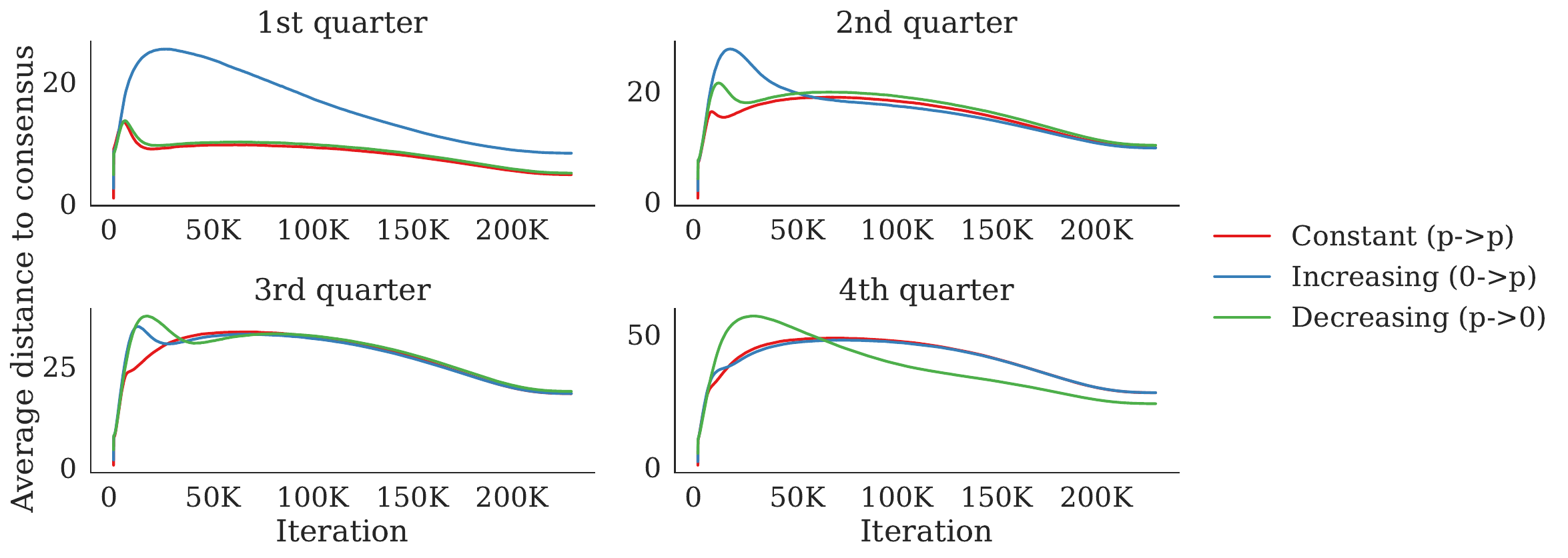}
    \caption{\textbf{Average distance to the consensus for different layer-wise adaptations of WASH}, for different slices of the model's parameters. Keeping the probability constant across layers ensures the lowest distance to consensus for the first quarters. Surprisingly, in the last quarter of parameters, despite initially starting with a higher distance to consensus, the `decreasing probability' shows a lower distance to consensus later in training; despite shuffling being less frequent than the other schedules. The `increasing probability' showcases how early layers are sensible to shuffling.}
    \label{fig:consensus_schedules}
\end{figure}

\begin{figure}[t]
\begin{subfigure}{0.48\linewidth}
    \centering
    \includegraphics[width=0.72\linewidth]{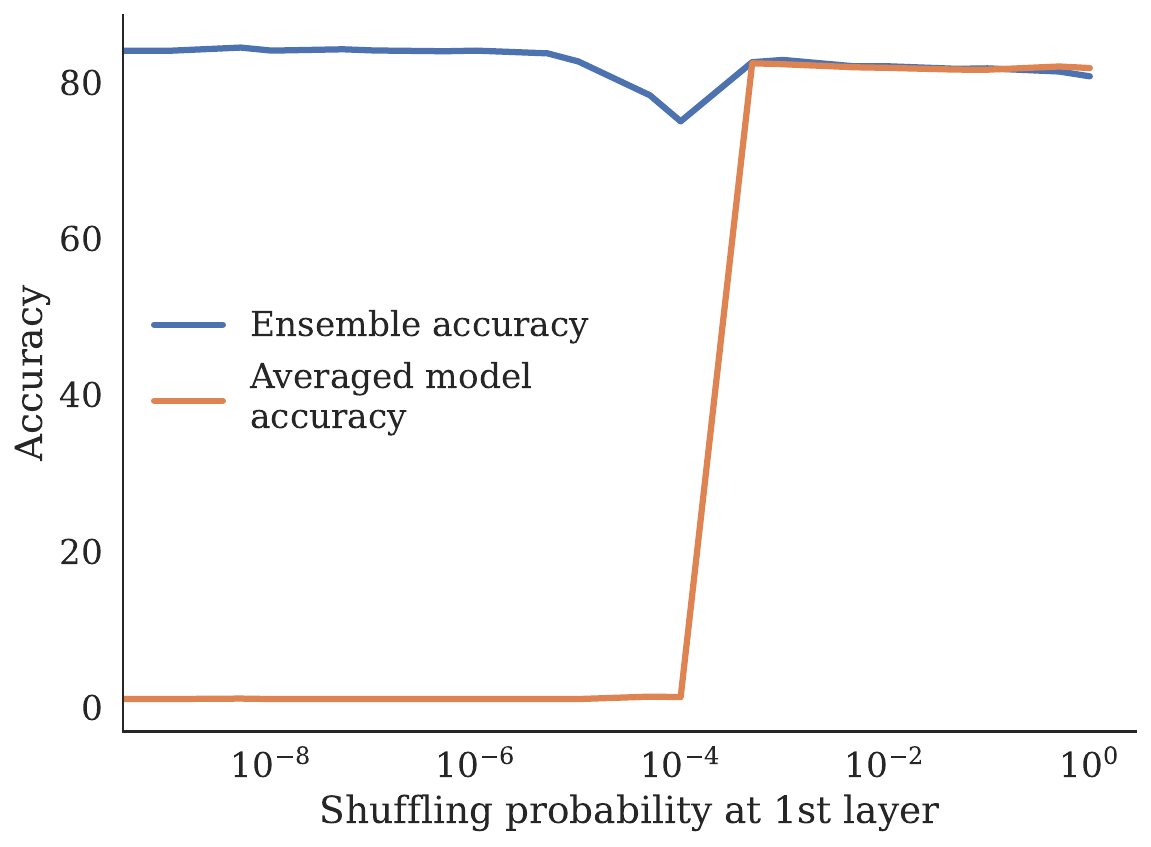}
    \caption{\textbf{Ensemble and Averaged accuracy for varying base probability values.} We observe a phase transition as the base probability augments between a phase where permuting does not improve the averaged model accuracy and a phase where the ensemble accuracy is equal to the averaged model accuracy. Between the phases, the ensemble accuracy diminishes.}
    \label{fig:acc_by_proba}
\end{subfigure}
 \hfill
 \begin{subfigure}{0.48\linewidth}
    \centering
     \centering
    \includegraphics[width=0.72\linewidth]{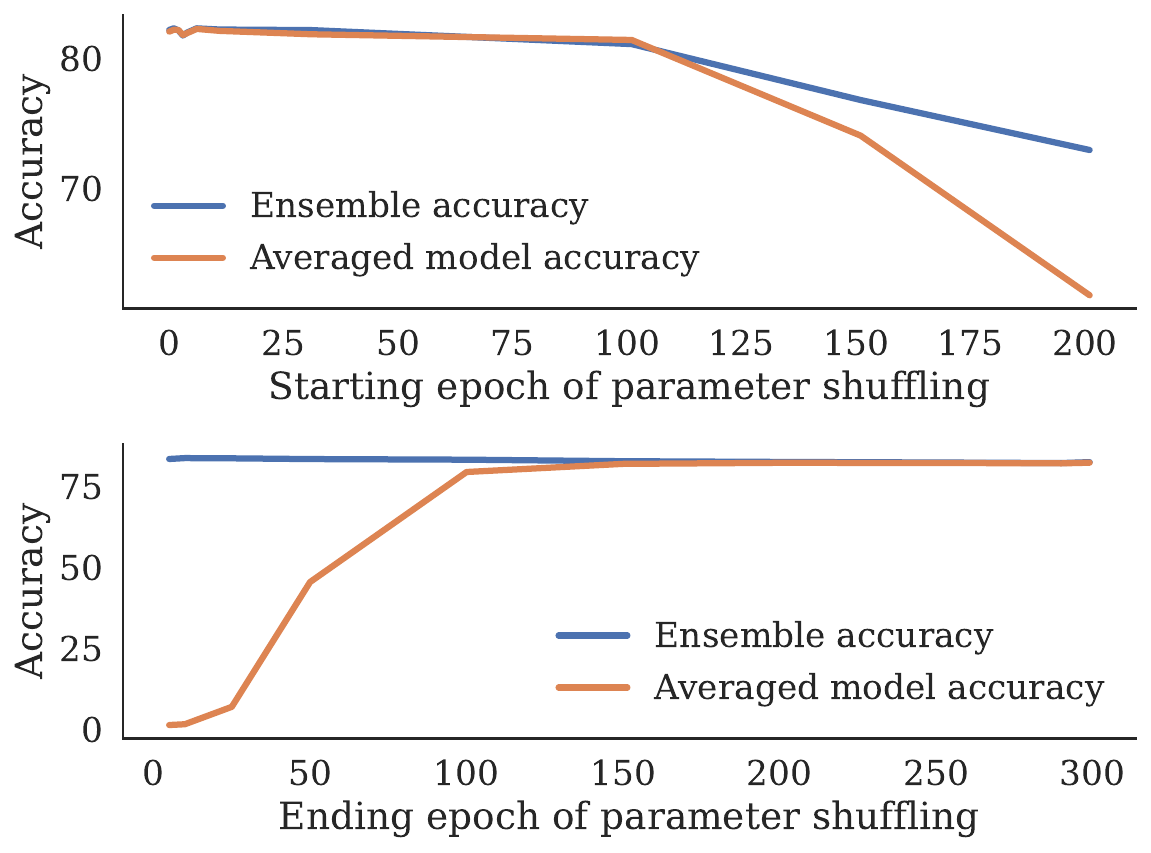}%
    \caption{\textbf{Ensemble and Averaged accuracy depending on the starting or ending epoch of the shuffling.} The parameter shuffling is beneficial both at the start and the end of training. Note that ending early, at epoch 150 out of 300, is less impactful on performances than starting permuting at epoch 150, showing that WASH is more important early in training. }
    \label{fig:acc_warmup}
\end{subfigure}
\caption{\textbf{Ablations of WASH}}
\end{figure}

\paragraph{Layer-wise adaptation variations.}

For WASH, we found that a probability decrease with depth provided the best results. We show in Tab. \ref{tab:schedules} of the Appendix the performances for alternatives, with probability either staying constant or increasing with depth. We find lower performances for both alternatives. 
In Fig. \ref{fig:consensus_schedules}, we report the models' distances to consensus for all three schedules. More precisely, we provide the distances for different slices of the models' parameters, indicating the effect of the shuffling depending on the depth. Shuffling equally all layers results in the lowest distance to the consensus as predicted, except in the last quarter of parameters. Here, surprisingly, our base `decreasing' schedule shows a lower distance to consensus despite shuffling less frequently. We also observe a particularly strong effect of the shuffling for early layers, as the distance is more emphasized in the first quarter between the `increasing' curve and the others. 

\paragraph{Base probability variation.} We present in Fig.~\ref{fig:acc_by_proba} the Ensemble and Averaged for different values of $p$, the base shuffling probability of the first layer. Rather than a smooth increase of the Averaged model accuracy, we observe a phase transition between a phase where the Averaged model accuracy is not improved by the shuffling and a sudden increase in the accuracy where it reaches the accuracy of the Ensemble. Just before the transition, the Ensemble accuracy is decreased, before increasing again back to its previous performance. The accuracy decreases only slightly even when increasing the shuffling probability to $1$, indicating the resilience of the models even to heavy shuffling. 

\paragraph{Shuffling is beneficial at every step.} Finally, we propose to show the impact of the parameter shuffling at different steps of the training by varying the epoch at which the shuffling either starts or stops. In Fig. \ref{fig:acc_warmup}, we show that there is no improvement by having a warmup or slowdown period in parameter shuffling, indicating that all phases of the training are improved by WASH. Furthermore, stopping parameter shuffling early results in a much smaller loss of Averaged accuracy compared to starting late. In other words, shuffling at the start of training before models converge is more impactful as models may still reside in different loss basins.


\section*{Conclusion}

We proposed a novel distributed training method, WASH, aimed at training a population of models in parallel. These models are averaged at the end of training to obtain a highly performing model with accuracies close to the ensemble accuracy for a fraction of the inference cost. Our method requires a fraction of the communication cost of similarly performing techniques while obtaining state-of-the-art results for our weight-averaged models. We show that our novel parameter shuffling does not explicitly reduce the distance between models while increasing the diversity of the optimization paths seen by the population. Still, we observe that the distance between our models is smaller than training them separately, allowing them to be averaged at the end of training.

\section*{Acknowledgements}

This work was supported by Project ANR-21-CE23-0030 ADONIS, EMERG-ADONIS from Alliance SU, and Sorbonne Center for Artificial Intelligence (SCAI) of Sorbonne University (IDEX SUPER 11-IDEX-0004). This work was granted access to the AI resources of IDRIS under the allocations 2023-A0151014526 made by GENCI. Eugene Belilovsky also acknowledges the FRQNT New Scholar grant.






\bibliographystyle{abbrv}
\bibliography{biblio.bib}

\section{Appendix}

\paragraph{2D optimization example}
The loss function we consider is a heavily simplified version of the Ackley function. With a minima in $(x_m,y_m)$ defined by 
\begin{equation}
    g(x,y,x_m,y_m, \lambda) = \exp{(-\lambda \sqrt{0.5 ((x-x_m)^2 + (y-y_m)^2})}\,,
\end{equation}
the function we consider in our example is 
\begin{equation}
f(x,y) = -10 g(x,y,10,10,0.1) - 5 g(x,y,8,3,0.3) - 5 g(x,y,3,8,0.3)\,.
\end{equation}
This function has a 2 local minima in $(3,8)$ and $(8,3)$ and a global minimum in $(10,10)$. In all three cases, the starting points are $(0,5)$ and $(5,0)$. We compute SGD by first computing the exact gradient of the function and then adding Gaussian noise to the gradient. The learning rate is $0.1$ and we optimize for $1000$ steps. For PAPA, we consider $\alpha = 0.99$. For WASH, the shuffling probability is equal for both coordinates and equal to $0.01$.

\paragraph{Interpolation heatmap}

\begin{figure}[t]
\begin{subfigure}{0.32\linewidth}
    \centering
    \includegraphics[width=0.95\linewidth]{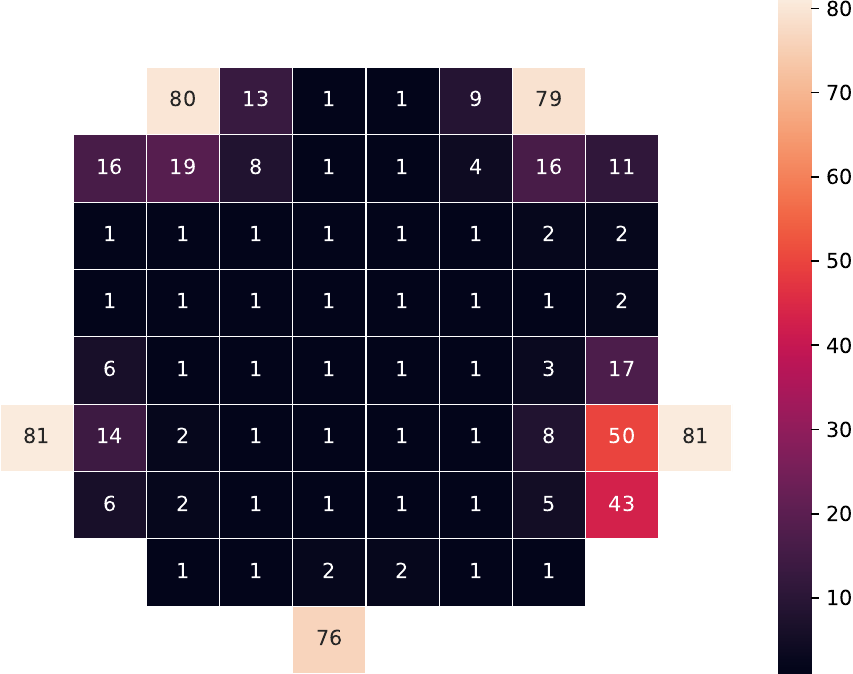}
    \caption{\textbf{Accuracy heatmap of the Baseline.} The interpolated models' performance is equal to random ones.}
    \label{fig:heatmap_base}
\end{subfigure}
 \hfill
 \begin{subfigure}{0.32\linewidth}
    \centering
     \centering
    \includegraphics[width=0.95\linewidth]{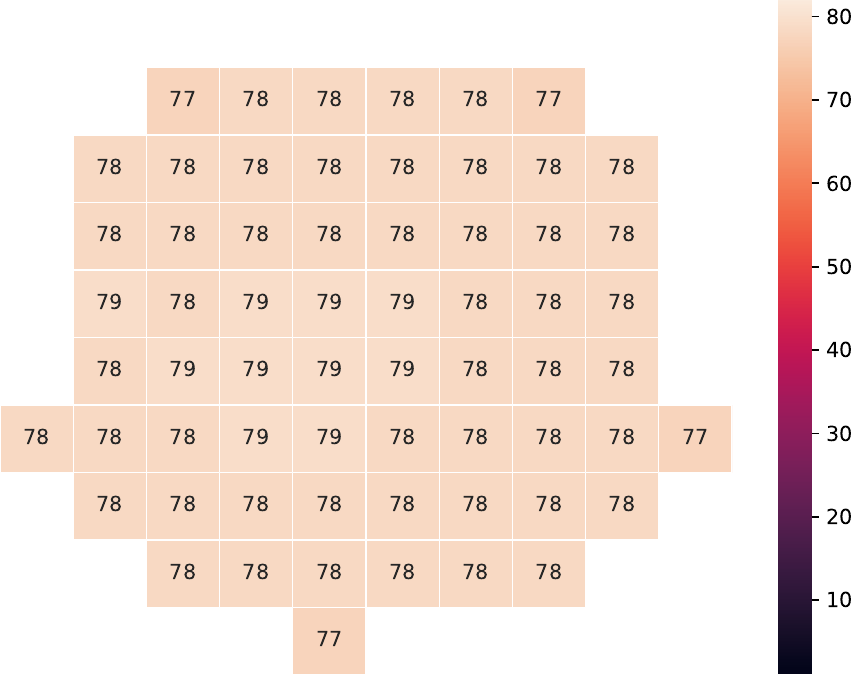}%
    \caption{\textbf{Accuracy heatmap of WASH.} The accuracy is very similar for various interpolations. }
    \label{fig:heatmap_wash}
\end{subfigure}
 \hfill
 \begin{subfigure}{0.32\linewidth}
    \centering
     \centering
    \includegraphics[width=0.95\linewidth]{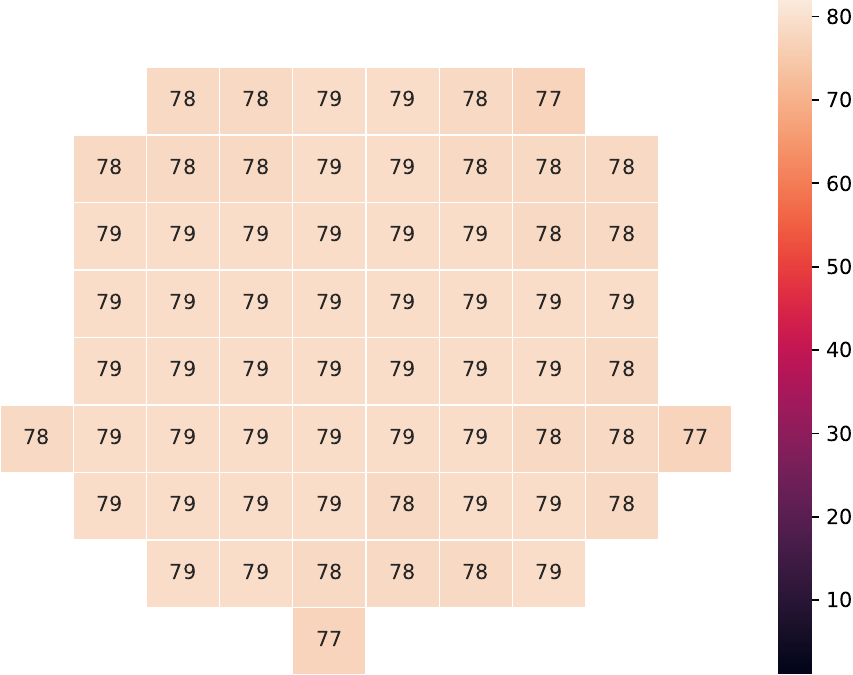}%
    \caption{\textbf{Accuracy heatmap of WASH+Opt.} The results are similar to WASH. }
    \label{fig:heatmap_washopt}
\end{subfigure}
\caption{\textbf{Accuracy heatmap for different weight interpolations}, for models trained separately, with WASH or WASH+Opt.}
    \label{fig:heatmaps}
\end{figure}

Here, we propose to display a heatmap showing the accuracy of more varied interpolations between $5$ models trained separately, with WASH, or WASH+Opt. We observe how WASH and WASH+Opt trained models converge to the same loss bassin, and that a large number of possible interpolations result in a high accuracy. The heatmaps are presented in Fig.~\ref{fig:heatmaps}.

\paragraph{Layer-wise adaptation variants performance} We showcase in Tab.\ref{tab:schedules} the performance of the three variants of layer-wise adaptations of WASH.

\begin{table}[t]
\centering
\caption{\textbf{Test accuracies of WASH with variants of the shuffling probability per depth.} Trained with a population of $5$ models on CIFAR-100 with a ResNet-18. The results show that permuting the first layers is more important than the later layers. Still, a constant probability across layers does not decrease WASH's performance much.}
\label{tab:schedules}
\resizebox{\textwidth}{!}{
\begin{tabular}{rll|lll|ll}
\toprule
\multicolumn{3}{c|}{\textbf{Proba. at layer}} & \multicolumn{3}{c|}{\textbf{Technique}} & & \\
    \textbf{0} & \textbf{to} & \textbf{L-1} & \textbf{Ensemble}  & \textbf{Averaged}   & \textbf{GreedySoup}  & \textbf{Best model} & \textbf{Worst model}    \\ \midrule
\textbf{$10^{-3}$}  & \textbf{$\searrow$}  &  \textbf{0}       & 82.22$\pm$.38 & 	82.15$\pm$.22 & 81.94$\pm$ 0.25 & 80.89$\pm$.03 &	78.80$\pm$.77 	 \\ 
   \textbf{$10^{-3}$}  & \textbf{$\rightarrow$}       &   \textbf{$10^{-3}$}  &  82.04$\pm$.19 & 	81.94$\pm$.15 & 81.69$\pm$.23 &	80.60$\pm$.16 &	78.67$\pm$.89   \\ 
  \textbf{0}  & \textbf{$\nearrow$}          &   \textbf{$10^{-3}$}  & 81.75$\pm$.35 & 	81.37$\pm$.10 & 81.14$\pm$.20 &	80.08$\pm$.40 & 78.55$\pm$.70 \\ 
   \bottomrule
\end{tabular}
}
\end{table}


\paragraph{Augmentations and regularization used} We follow the same data augmentations and regularizations used in \cite{2304.03094_papa} for a fair comparison. We use Mixup (random draw from \{0, 0.5, 1.0\} for CIFAR-10/100 or from \{0, 0.2\} for ImageNet), Label smoothing (random draw from \{0, 0.05, 0.1\} for CIFAR-10/100 or from \{0, 0.1\} for ImageNet), CutMix (random draw from \{0, 0.5, 1.0\} for CIFAR-10/100 or from \{0, 1.0\} for ImageNet) and Random Erasing (random draw from \{0, 0.15, 0.35\} for CIFAR-10/100 or from \{0, 0.35\} for ImageNet).

For our experiments, we required a single A100 GPU for up to 14 hours to train up to a population of 10 models, and up to 40 hours for a population of 20 models. Similarly, we required 16 A100 GPUs to train in parallel a population of 5 models on ImageNet.

\end{document}